%% file: main.tex
\documentclass[sigconf,authorversion]{acmart}

\AtBeginDocument{%
  \providecommand\BibTeX{{%
    \normalfont B\kern-0.5em{\scshape i\kern-0.25em b}\kern-0.8em\TeX}}}

\copyrightyear{2023} 
\acmYear{2023} 
\setcopyright{acmlicensed}\acmConference[ETRA '23]{2023 Symposium on Eye Tracking Research and Applications}{May 30-June 2, 2023}{Tubingen, Germany}
\acmBooktitle{2023 Symposium on Eye Tracking Research and Applications (ETRA '23), May 30-June 2, 2023, Tubingen, Germany}
\acmPrice{15.00}
\acmDOI{10.1145/3588015.3588416}
\acmISBN{979-8-4007-0150-4/23/05}

\acmSubmissionID{1038}

\usepackage{enumitem}
\usepackage{textcomp}

\begin{document}

\title{Introducing Explicit Gaze Constraints to Face Swapping}

\author{Ethan Wilson}
\email{ethanwilson@ufl.edu}
\orcid{0000-0003-0944-2641}
\affiliation{
  \institution{University of Florida}
  \city{Gainesville}
  \state{Florida}
  \country{USA}
}

\author{Frederick Shic}
\email{fshic@uw.edu}
\orcid{0000-0002-9040-1259}
\affiliation{
  \institution{University of Washington}
  \city{Seattle}
  \state{Washington}
  \country{USA}
}

\author{Eakta Jain}
\email{ejain@ufl.edu}
\orcid{0000-0001-5131-3355}
\affiliation{
  \institution{University of Florida}
  \city{Gainesville}
  \state{Florida}
  \country{USA}
}

\input{abstract.tex}




\maketitle

\input{introduction.tex}
\input{related_work.tex}
\input{methodology.tex}
\input{results.tex}
\input{discussion.tex}


\bibliographystyle{ACM-Reference-Format}
\bibliography{bibliography}

\input{appendix.tex}

\end{document}

%% file: abstract.tex
\begin{abstract}

Face swapping combines one face's identity with another face's non-appearance attributes (expression, head pose, lighting) to generate a synthetic face.  This technology is rapidly improving, but falls flat when reconstructing some attributes, particularly gaze.  Image-based loss metrics that consider the full face do not effectively capture the perceptually important, yet spatially small, eye regions.  Improving gaze in face swaps can improve naturalness and realism, benefiting applications in entertainment, human computer interaction, and more.  Improved gaze will also directly improve Deepfake detection efforts, serving as ideal training data for classifiers that rely on gaze for classification.  We propose a novel loss function that leverages gaze prediction to inform the face swap model during training and compare against existing methods.  We find all methods to significantly benefit gaze in resulting face swaps.  
    
\end{abstract}

%% file: introduction.tex
\section{Introduction}

Face swapping is the act of placing a \textit{character's} face overtop of an \textit{original} face in a piece of media.  
In deep learning, face swapping is distinct from the face generation task 
--- the network needs to create a realistic face that preserves the original attributes (such as head pose, gaze direction, and mouth movements) while having explicit control over the face's identity.  Current face swapping methods have solved the identity reconstruction task, but generally match all other attributes of the face using a black-box approach.

We leverage a pretrained gaze estimation network to optimize an existing face swapping pipeline.  Using predicted gaze angles of original and reconstructed faces, we define a reconstruction loss term focused on the eyes to add a gaze component to the overall optimization function, enhancing the accuracy of reconstructed gaze without compromising visual fidelity.  Our explicit focus on preserving gaze behavior could be applied to future face swapping pipelines.  Our implementation alters the optimization function but does not alter model architecture, meaning that already-trained models can be fine-tuned with this improvement in place.  We showcase the method on a popular open-source face swapping network, seeing significant improvement in reconstructed gaze directions compared to baseline face swapping.

\subsection{Main Contribution}

The proposed design improves upon the naturalness and correctness of gaze behavior in generated face swaps, incorporating a pretrained deep-learning network to guide training in a novel way.  Our methodology and experiments provide an implementation guide for facial attribute-based loss functions and reveal their effectiveness, respectively.  


\subsection{Ethics of Face Swapping}

Face swapping has uses in visual effects, interactions with virtual avatars~\cite{caporusso_deepfakes_2021, foreman_salvador_2019}, and privacy protection~\cite{zhu_deepfakes_2020, lee_american_2021, wilson_practical_2022}; however, face swapping has become a controversial technology due to its potential for impersonation, spreading misinformation and violating individuals' privacy.  These so called \textit{Deepfakes}' sudden accessibility has incited public concern and sparked legislative response~\cite{wagner_real_2019}.  Yet, responsible innovation on face swapping is necessary and will lead to positive outcomes.  The methods this paper explores can increase the naturalness of future face swapping algorithms, making them more feasible in positive applications for social good.  These innovations will also aid in the detection of Deepfakes.  Classifiers based on biometric signals, including gaze patterns, are being developed for Deepfake detection~\cite{li_ictu_2018, jung_deepvision_2020, ciftci_fakecatcher_2020, ciftci_heartbeat_2020, demir_wherelook_2021}.  These methods train on real and swapped face videos.  By feeding these models new training data with more believable gaze, we will see increased accuracy and reliability when detecting fake media across the internet.

%% file: related_work.tex
\section{Related Work}

Recent innovations in image generation techniques, most prominently the generative adversarial network (GAN)~\cite{goodfellow_generative_2014}, variational autoencoder (VAE)~\cite{kingma_autoencoding_2013} and improvements thereafter~\cite{karras_style-based_2019, karras_analyzing_2020, razavi_generating_2019, radford_unsupervised_2016, liu_coupled_2016, hou_deep_2017}, have rapidly advanced the ability to create realistic AI-synthesized faces.  These technologies paved the way for powerful, fully automated face swaps that have become nearly undetectable to naive human viewers.  

The original image-based face swapping algorithm~\cite{deepfakes_faceswap_2022} is a forked autoencoder with two distinct decoders, each training on a unique identity.  Advancements over this initial method have focused on swapping between arbitrary identities~\cite{nirkin_fsgan_2019, li_faceshifter_2019, chen_simswap_2020}, real-time applications~\cite{korshunova_fast_2017}, and achieving higher resolutions~\cite{naruniec_neural_2020, zhu_oneshot_2021}.  

Some image and face synthesis methods have begun to leverage existing networks, hereafter referred to as \textit{pretrained expert models}, as part of their training process.  Facial recognition systems~\cite{deng_arcface_2019} have been used to automatically segment identity or to obtain an overall attribute profile~\cite{korshunova_fast_2017, chen_simswap_2020, tang_cycle_2019, nitzan_face_2020}; facial attribute extractors have been used to classify the face in an unsupervised manner~\cite{li_anonymousnet_2019}; landmark estimators have been used to extract or enforce body/facial structure~\cite{sun_natural_2018, nitzan_face_2020, kuang_effective_2021, siarohin_motion_2021}; style transfer algorithms extract style using pretrained networks' intermediate features~\cite{liu_blendgan_2021, gatys_style_2016, johnson_perceptual_2016, zhang_multistyle_2018}.  These methods found success using high-level predictions from pretrained expert models to aid in training without requiring supervised labels. 

The core goal of modern face swapping is to disentangle the embedded feature vector between identity and other facial attributes, so that identities can be swapped while all other features remain constant.  While each algorithm is unique, in nearly all methods the problem is framed as \textbf{identity} \textit{versus} \textbf{all attributes}, i.e. all aspects outside of identity are placed under a single loss term.  For example, multiple approaches isolate and replace the identity portion within an autoencoder's feature embedding~\cite{korshunova_fast_2017, chen_simswap_2020, li_faceshifter_2019, wang_hififace_2021}. The overall attribute profile is preserved, but is enforced only according to a general image-based reconstruction loss, which may fail to emphasize perceptually relevant features.  Particularly, the eyes spatially occupy only about 5.6\% of the face, yet human viewers focus on the eyes approximately 40\% of the time~\cite{janik_eyefocus_1978}.  Because features are derived implicitly from pixel images, the eyes are not prioritized, thus have been found to account for a large percent of noticed artifacts~\cite{wohler_towards_2021}.  

A simple way to improve results is the brute-force approach --- create a deeper network with a larger latent space.  For example, using eight identity-specific decoders rather than two and increasing model depth~\cite{naruniec_neural_2020}.  This is effective yet sees increased training times and memory requirements.  Instead of increasing resources, another potential solution is to add a gaze-aware constraint to the training process, but this method is not well explained or evaluated in the corresponding manuscript~\cite{perov_deepfacelab_2021}.
 
This work details methods to impose explicit constraints on the facial attribute profile, incentivizing the network to better preserve the behavior of the eyes.  Our proposed method is modular and could easily extend existing face swapping architectures, leveraging pretrained expert models to better inform models of perceptually important features such as gaze.

%% file: methodology.tex
\section{Methodology}

We propose a novel method to explicitly prioritize gaze over all other implicitly defined facial attributes when training face swapping models.  The proposed method leverages a pretrained gaze estimation network, using the resulting gaze values to formulate a reconstruction loss focused on the eye region of the face.  Our approach is flexible and generalizes, meaning that it can be applied to any face swapping architecture and with other pretrained expert models.  Already-trained models could also be fine-tuned with this improvement.  We evaluate our method on DeepFaceLab (DFL)~\cite{perov_deepfacelab_2021}, comparing against both a gaze-unaware baseline model and their native solution, which had not been formally analyzed or explained in the literature.

\subsection{Overview of DeepFaceLab}

DFL is the most popular publicly available face swapping platform, so is representative of the majority of face swaps found online.  There are many resources online to aid in understanding DFL's pipeline\footnote{https://mrdeepfakes.com/forums/thread-guide-deepfacelab-2-0-guide}. 
 For explanation and justification of DFL's model design, please refer to their manuscript\footnote{https://arxiv.org/abs/2005.05535}.

\begin{figure}[h]
    \centering
    \includegraphics[width=1\linewidth]{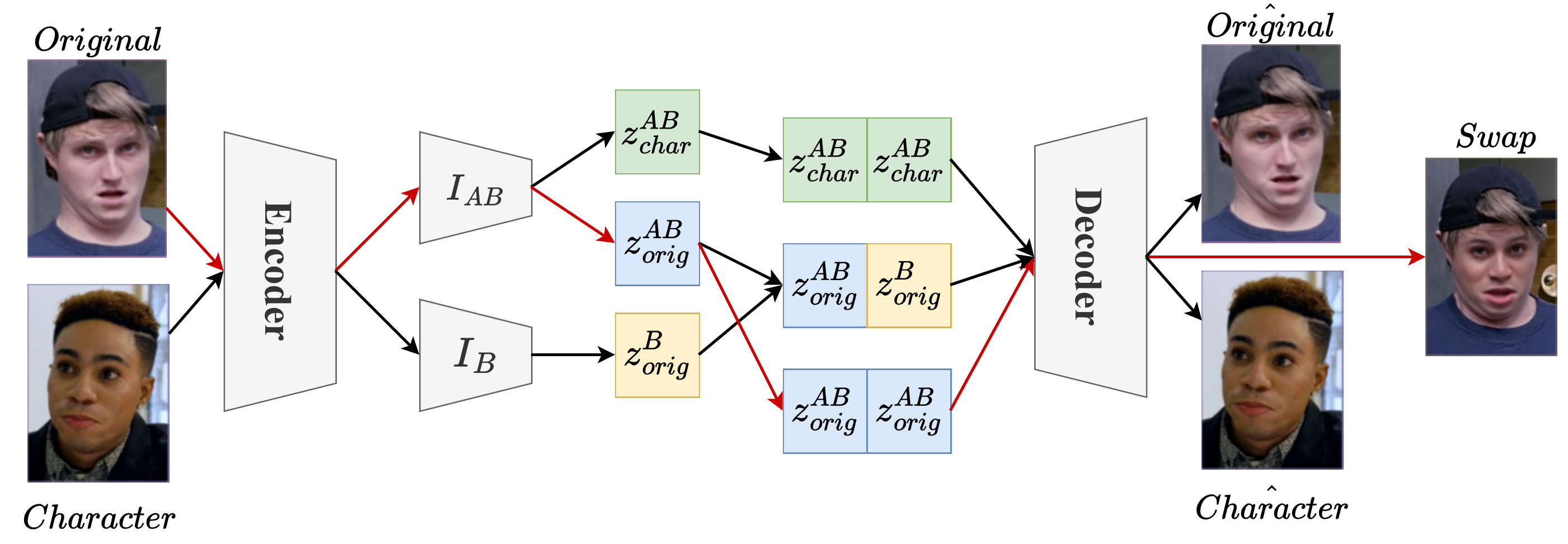}
    \caption{Illustration of DFL's LIAE architecture.  The pathway taken to create the resulting face swap is displayed in red.  Note that $z^{AB}_{char}$ is concatenated with a copy of itself to reconstruct the character face, and $z^{AB}_{orig}$ is concatenated with itself to produce the face swap result.}
    \label{fig:liae_design}
\end{figure}

We use DFL's LIAE architecture, which disentangles identity with intermediate networks between the encoder and decoder (see Figure~\ref{fig:liae_design}).  The first intermediate network $I_{AB}$ generates latent vectors $z^{AB}_{char}$ and $z^{AB}_{orig}$ during the training process.  The second intermediate network $I_B$ is only given the original identity to generate $z^{B}_{orig}$.  Before passing to the decoder, the latent vectors are concatenated: the original face's becomes $z^{AB}_{orig} || z^{B}_{orig}$ and the character face's concatenates a copy of itself to become $z^{AB}_{char} || z^{AB}_{char}$.  These latent vectors are passed through the respective decoders to reconstruct the input faces.  During face swapping, the original face is only passed through $I_{AB}$ and concatenated onto itself to generate latent code $z^{AB}_{orig} || z^{AB}_{orig}$, which is then fed through the decoder to generate a face swapped result.  

Intuitively, we can interpret the first latent vector $z_1$ to contain attributes, the second vector $z_2$ to contain identity information, and $z_1 || z_2$ to contain full facial information.  The latent vector $z^{AB}$ never represents the original face's identity during training, so becomes hardwired to the character face's identity.  The LIAE design can be seen in Figure~\ref{fig:liae_design}.  

During training, DFL uses segmentation masks to isolate the error calculation to relevant parts of the face~\cite{bulat_2017}.  The three masks utilized are of the face ($M_{face}$), the eyes ($M_{eyes}$), and the eyes plus mouth ($M_{em}$).  In the following equations, we define the input faces as $Y$ and their reconstructions as $\hat{Y}$\footnote{Note that original and character faces' reconstruction loss are computed in identical fashion.}.  The reconstruction loss combines difference of structural similarity (DSSIM)~\cite{wang_ssim_2004, zhao_ssim_2017} and mean squared error (MSE).  DSSIM enforces structural consistency between the input and output face using luminance, contrast, and structural components, and MSE error enforces pixel-wise similarity.  The loss equations are as follows:





{
\begin{gather}
    SSIM(Y, \hat{Y}) = \frac{(2\mu_i\mu_j + c_1)(2\sigma_{ij} + c_2)}{(\mu_i^2 + \mu_j^2 + c_1)(\sigma_i^2 + \sigma_j^2 + c_2)} \\ 
    L_{DSSIM}(Y, \hat{Y}) = \frac{1 - SSIM(Y, \hat{Y})}{2}
\end{gather}

\centering{where $i$, $j$ $=$ sliding windows of size $N$x$N$ 

$\mu_i$, $\mu_j$ $=$ average of $i$, $j$ $\qquad$ $\sigma_i^2$, $\sigma_j^2$ $=$ variance of $i$, $j$ $\qquad$ 

$\sigma_{ij}$ $=$ covariance of $i$, $j$ $\qquad$ $c_1$, $c_2$ $=$ stabilizing variables}

}

\begin{equation}
    L_{MSE}(Y, \hat{Y}) = \frac{1}{n} \sum_{i=1}^{n} (\hat{Y_i} - Y_i)^2
\end{equation}

The core reconstruction loss is a weighted sum between DSSIM, MSE, and an MSE calculation comparing the input and predicted face masks:

\begin{multline}
    \label{eqn:core_loss}
    L_{\triangle}(Y, \hat{Y}, M_{face}, \hat{M_{face}}) = \lambda_1 L_{DSSIM}(Y, \hat{Y}) + \\ 
    \lambda_2 L_{MSE}(Y, \hat{Y}) + \lambda_3 L_{MSE}(M_{face}, \hat{M_{face}})
\end{multline}

DFL can explicitly target facial attributes via its optional eyes and mouth priority term.  This integrates well with the main loss equation, measuring the absolute value of pixel error between the original and generated faces masked to the eyes and mouth.  This is an optional term that must be enabled by DFL users.

\begin{equation}
        L_{\triangle em}(Y, \hat{Y}, M_{em}) = \lambda_{em} |Y M_{em} - \hat{Y} M_{em}|
\end{equation}

\subsection{Proposed Gaze Reconstruction Loss}

Motivated by previous image generation methods' success using pretrained expert models, we leverage a gaze estimation network.  We incorporate L2CS-Net\footnote{https://github.com/Ahmednull/L2CS-Net}~\cite{abdelrahman_l2cs_2022}, which predicts pitch and yaw angles $\mu$, $\phi$ from input face images.  This network is optimized towards unconstrained environments so is well suited to the data typical in training face swaps.   We incentivize the face swapping model to better reconstruct gaze by penalizing offsets in predicted gaze angle between the input and reconstructed faces during training.

Our gaze reconstruction loss is computed as follows.  $\mu$ and $\phi$ are converted to normalized Cartesian coordinates, then the angle $\theta$ between the two vectors is found.
\begin{gather*}
    \mu, \phi = L2CS(Y) \qquad \qquad \hat{\mu}, \hat{\phi} = L2CS(\hat{Y}) \\
    x = sin(\phi) cos(\mu) \qquad \qquad y = sin(\phi) sin(\mu) \\ 
    z = cos(\phi) \qquad \qquad \qquad V_1 = <x, y, z> \\
    \hat{x} = sin(\hat{\phi}) cos(\hat{\mu}) \qquad \qquad \hat{y} = sin(\hat{\phi}) sin(\hat{\mu}) \\ 
    \hat{z} = cos(\hat{\phi}) \qquad \qquad \qquad V_2 = <\hat{x}, \hat{y}, \hat{z}>
\end{gather*}

\begin{equation}
    \textrm{Error is computed as:} \qquad \theta(Y, \hat{Y}) = cos^{-1}\Biggl(\frac{V_1 \cdot V_2}{\|V_1\| \|V_2\|}\Biggr)
\end{equation}

\begin{figure}[t]
    \centering
    \includegraphics[width=1\linewidth]{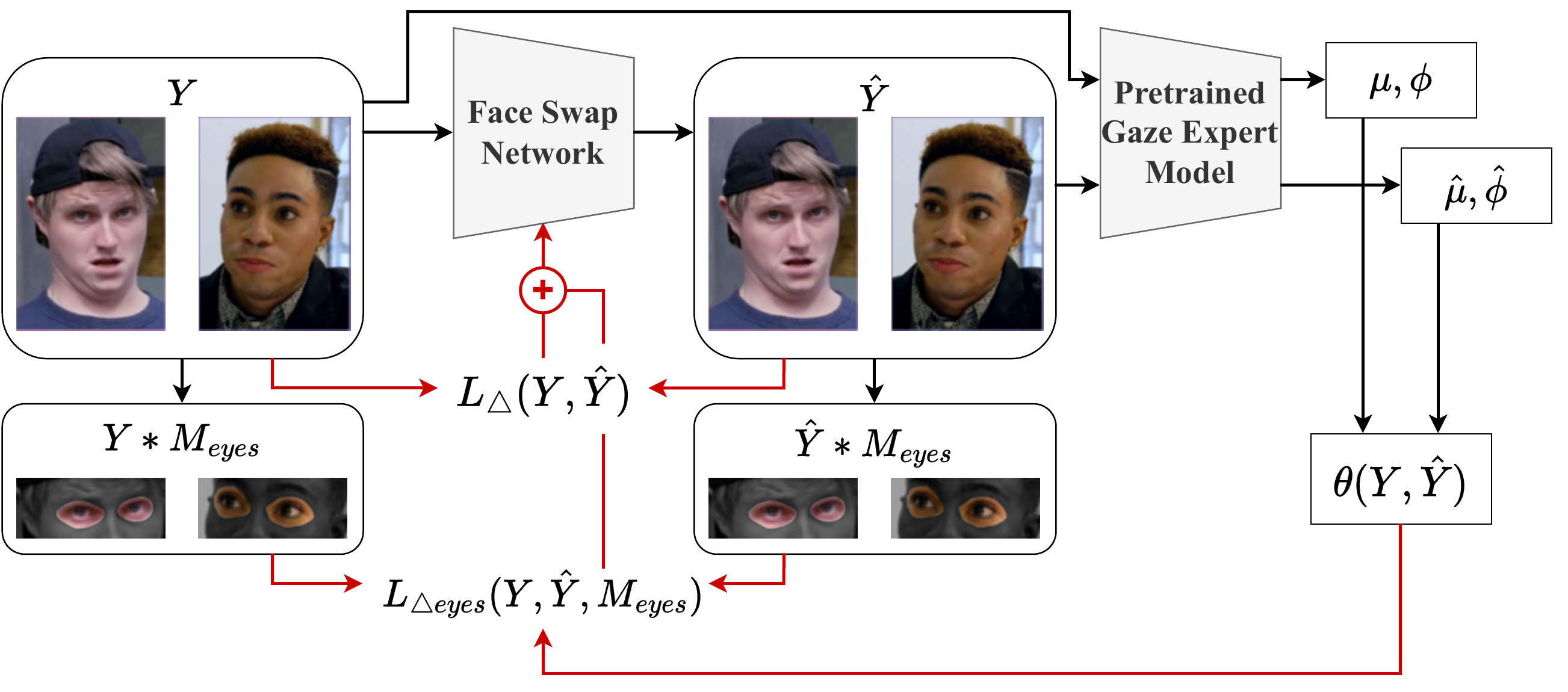}
    \caption{Design diagram of the steps to compute the gaze reconstruction loss.}
    \label{fig:gaze_pipeline}
\end{figure}

We apply this error term only to the regions of the network that correspond to the eyes.  We use $Y$, $\hat{Y}$, and $M_{eyes}$ to construct a reconstruction loss specific to the eyes that can be scaled by the computed $\theta$ and hyperparameters $\alpha$ and $\beta$.  We structure our loss equation similarly to equation~\ref{eqn:core_loss}, using DSSIM and MSE computations on the original and reconstructed image eye regions.  An illustration of the design and steps taken to compute the loss equation can be seen in Figure~\ref{fig:gaze_pipeline}.

\begin{multline}
    L_{\triangle eyes}(Y, \hat{Y}, M_{eyes}) = \theta(Y, \hat{Y}) \Bigl(\alpha L_{DSSIM}(Y M_{eyes}, \hat{Y} M_{eyes}) + \\
    \beta L_{MSE}(Y M_{eyes}, \hat{Y} M_{eyes})\Bigr)
\end{multline}



%% file: results.tex
\section{Evaluation}

We assess the performance of each condition by analyzing the offset in viewing angles between the face swap and the real face in the corresponding source video.  To compute this metric, we utilize L2CS-Net~\cite{abdelrahman_l2cs_2022} to predict a gaze viewing angle for each condition, considering the source video's predicted gaze vector to be the ground truth.  In our evaluation we use DFL's set parameters for our $\lambda$ values.  Namely, $\lambda_1, \lambda_2, \lambda_3 = 10$, $\lambda_em = 300$.  When implementing our proposed loss term, we use $\alpha = 3$ and $\beta = 30$.

\subsection{Dataset}

We introduce a dataset to serve as a testing ground for our approach.  We generate our face swaps using the source video clips taken from the FaceForensics++ Deep Fake Detection Dataset\footnote{https://ai.googleblog.com/2019/09/contributing-data-to-deepfake-detection.html}~\cite{rossler_ff++_2019}.  In the dataset, subjects perform the same tasks\footnote{The video segments we use are: exit phone room, kitchen pan, outside talking pan laughing, walking outside cafe disgusted.  These are concatenated into a single video \~2 minutes in length per subject.}, ensuring similar expression and head pose, making these clips ideal for high quality face swaps.  Our dataset consists of six subject (three female, three male).  For each gender, two subjects have similar appearance to one another.  Per gender, we permute all combinations of subjects being used as the character and original face, resulting in a total of 12 unique face pairs. 


\begin{figure}[h]
    \centering
    \includegraphics[width=1\linewidth]{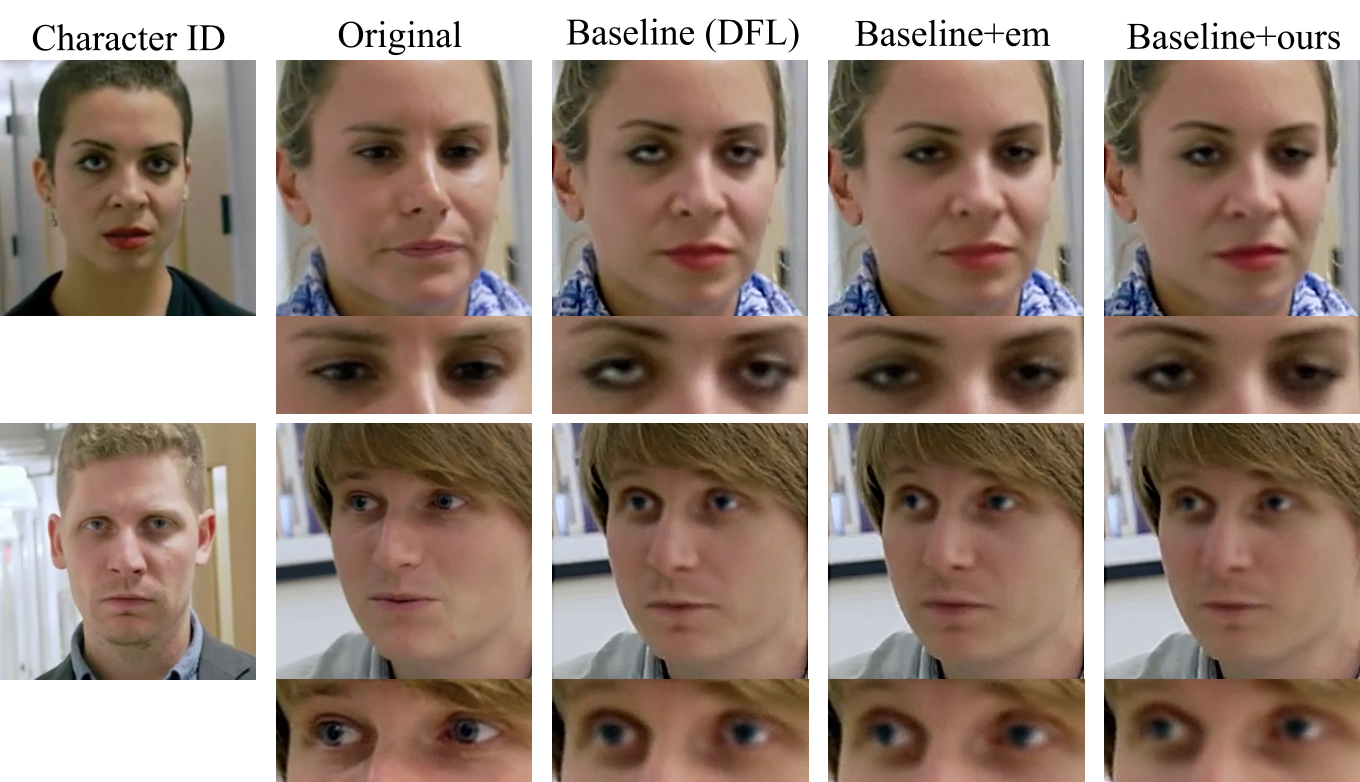}
    \caption{Visual comparison of face swaps produced by the baseline DFL method, DFL with eyes and mouth priority loss (em), and DFL with our proposed gaze loss.  Both improvements over the baseline reduce gaze angle error.}
    \label{fig:faces_figure}
\end{figure}

We generate face swaps across multiple conditions, keeping all other hyperparameters consistent.  
Every model is pretrained for 100 thousand iterations on the CelebA dataset~\cite{liu_faceattributes_2015}, then trained for the final 20 thousand iterations on the identity pair.  Frames from our generated dataset can be seen in Figure~\ref{fig:faces_figure}.  The conditions are:

\begin{itemize}
    \item \textbf{DFL.}  The model implicitly learns gaze behavior while optimizing the core reconstruction loss in equation~\ref{eqn:core_loss}.

    \item \textbf{DFL+em.}  DeepFaceLab with eyes and mouth priority loss enabled.  DFL's native solution which further enforces pixel-wise similarity for the key regions of the face.

    \item \textbf{DFL+Gaze.}  DeepFaceLab with our proposed gaze loss.  The model explicitly enforces consistency using gaze vectors computed by the pretrained expert model.

    \item \textbf{DFL+Gaze (finetuning).}  The model is pretrained with no gaze-specific loss, then trained for the final 20 thousand iterations using our proposed loss.

    \item \textbf{DFL+em+Gaze.}  Both DFL's native approach and our proposed approach are enabled during training.
\end{itemize}

\subsection{Results}

We analyze error values, collapsing from individual frames ($\sim2900$ per video) to average across each individual in the dataset.  The baseline DFL produces an average error of 5.98\textdegree [95\% Confidence Interval (CI): 4.82, 7.13].  All improvements on the baseline method produce noticeably more accurate gaze values:  DFL+em averages 4.85\textdegree [95\% CI: 3.80, 5.90], DFL+Gaze averages 4.71\textdegree [95\% CI: 3.66, 5.77], DFL+Gaze (finetuning) averages 4.85\textdegree [95\% CI: 3.80, 5.90], and DFL+em+Gaze averages 4.72\textdegree [95\% CI: 3.67, 5.77].  
On the test dataset, introducing DFL's eyes and mouth priority term decreases reconstructed gaze error by 18.1\%; introducing the proposed method decreases by 19.7\%, and introducing both components decreases gaze error by 20.32\%.

\begin{figure}[h]
    \centering
    \includegraphics[width=1\linewidth]{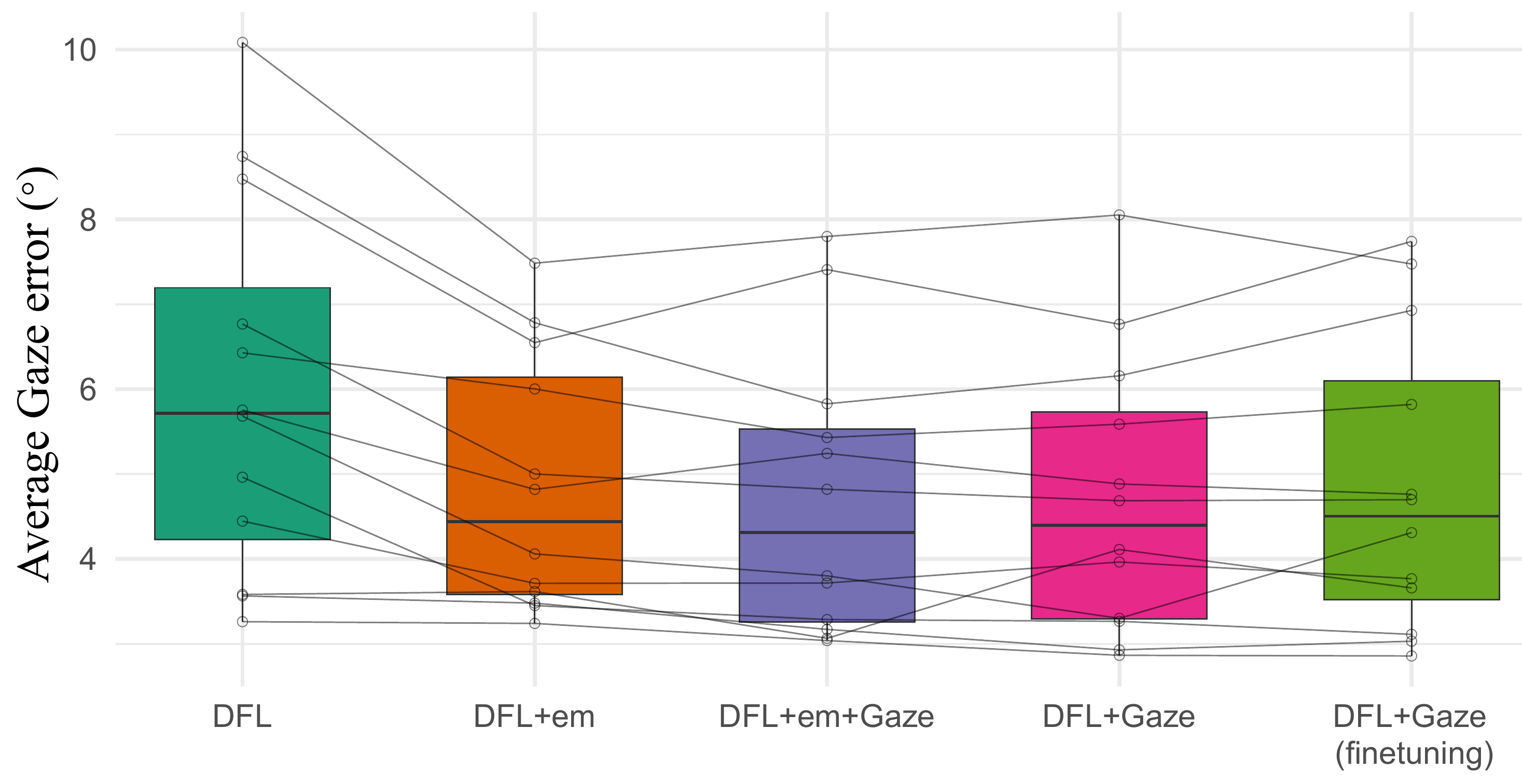}
    \caption{Plot of mean gaze error across all evaluated videos ($N=12$) by condition.  Individual video results are plotted over-top and connected across each box plot.}
    \label{fig:box_whisker}
\end{figure}

We test for significance via a linear mixed-effects model.  We first compute the average of the log of angular error for each method and individual, applying the log transform to improve normality of error distributions. We then model errors as average(log(error)) ~ method with a random intercept per individual.  All improved methods significantly improve over DFL ($p < 0.001$).  However, we have not found statistical evidence pair-wise between any of the improved methods.  Interestingly, the DFL+em+Gaze approach combining pixel information and explicit gaze modeling yielded insignificant benefit over DFL+em (t$(1,44)=1.603, p=0.116$).  This may indicate that the two optimizations capture similar underlying information.

\begin{figure}[ht]
    \centering
    \includegraphics[width=1\linewidth]{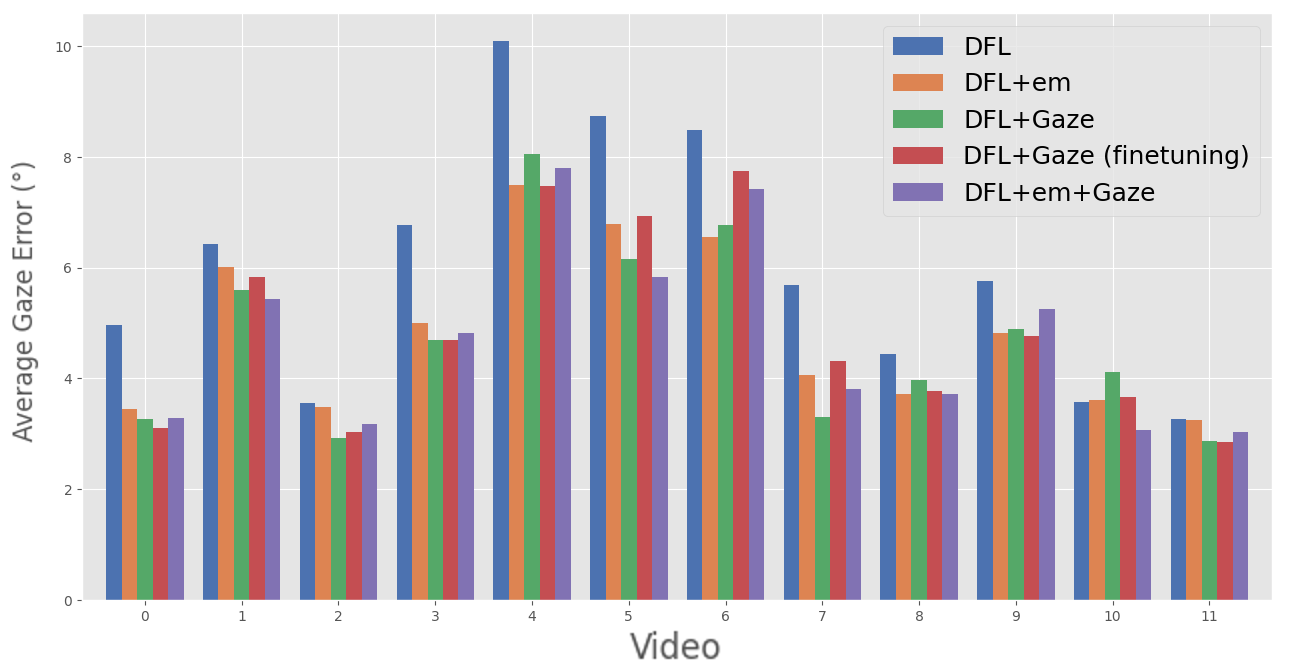}
    \caption{Average gaze error in degrees for all conditions evaluated, plotted by individual video.}
    \label{fig:bar_graph}
\end{figure}

Each method's performance across individuals in the dataset is plotted in Figure~\ref{fig:box_whisker}.  We see a large amount of variability among individual video results in all methods other than DFL, indicating roughly equivalent performance for all improvements analyzed.  Looking on an individual video basis (Figure~\ref{fig:bar_graph}), relative error remains quite stable across the dataset, suggesting that error is tied to the properties of the video, i.e. the specific pair of faces involved.

%% file: discussion.tex
\section{Discussion \& Conclusion}

Based on our experiments, the proposed gaze improvement for face swapping using a pretrained gaze prediction model largely decreases gaze error by 19.7\% when appended to an image-based reconstruction loss equation.  This analysis provided key information in enhancing gaze behavior in face swapping models and the potential benefit that can be provided.

Our method achieves similar performance compared to DFL's native solution to the problem (which had not previously been quantified relative to the baseline).  These adjunct approaches likely capture the same information.  However, it is important to note that the vast majority of face swapping approaches implement \textbf{neither} approach, so either will improve gaze representation.  Compared to the baseline, a few degrees may or may not have a noticeable impact on viewer perception.  However, these improvements could be quite beneficial to aid in the development of biometric Deepfake classifiers that leverage gaze to label video as real or fake.

Our method uses the same pretrained network in training as our evaluation pipeline.  This opens up the possibility that our model could have learned to minimize the prediction error for L2CS-Net rather than generally improving gaze representations.  However, by observing how visually similar our results are to the native solution and the minimal differences in gaze errors, this concern is alleviated.  Our pipeline leveraged the pretrained gaze model to derive an angle error $\theta$.  If we had instead granted white-box access to the pretrained model, fitting to the gaze model would be more likely.

Unlike the native approach, our proposed method incorporates gaze angle as a high-level feature.  This lessens the dependence on pixel-level matching of the eyes, possibly being more impactful at higher resolutions.  While this analysis focused fully on gaze, a similar pipeline could be easily developed for other features, such as expression or head-pose matching.  Stacking multiple optimizations on the same network could improve overall fidelity.  Analyzing the interaction between multiple pretrained expert models as they guide the same model's training process is a worthwhile future direction.  The dependence on pretrained expert models to compute gaze vectors will make our approach more appealing as more advanced predictors are developed.  For example, current gaze predictors are prone to around 4 degrees of prediction error, which is likely acting as a lower bound on our method's performance.  When better performing predictors are created, our system will improve accordingly.  

In this paper, we presented a novel loss component that significantly increases a face swapping model's ability to accurately reconstruct gaze.  We compared multiple design decisions, including a formal analysis of DFL's eyes and mouth priority method.  Our most successful implementation, combining both optimizations, decreased gaze reconstruction error by 20.32\%.  This advancement improves face swapping technology but is particularly promising for gaze-based Deepfake detection; such an increase in fidelity will allow researchers to generate higher quality training datasets that will lead to better Deepfake detection in real-world settings.

%% file: appendix.tex
\newpage
\appendix

\section{Supplementary Information}

\subsection{DFL Parameters}

\begin{figure}[h]
    \centering
    \includegraphics[width=1\linewidth]{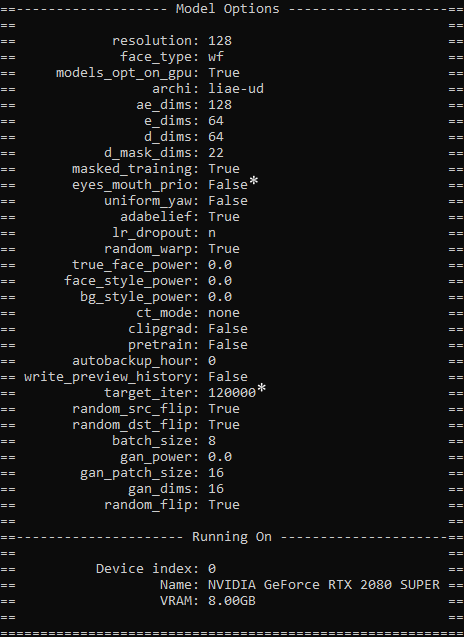}
    \caption{Training parameters used to when generating all face swap stimuli.  Note that the LIAE architecture is classified as a SAEHD model in DFL's configurations.  The \textbf{eyes\_mouth\_prio} parameter was set to true when the proposed gaze term was disabled.  \textbf{target\_iter} varied depending on training phase (100k iterations pretraining, 20k iterations on the end pair of identities).}
    \label{fig:train_params}
\end{figure}

All face swaps were generated using a NVIDIA GeForce RTX 2080 Super.  The exact run configurations used to generate our face swaps are outlaid.  For training parameters, refer to Figure~\ref{fig:train_params}.  For merge parameters, refer to Table~\ref{tab:merge_params}.

\begin{table}[h]
\centering
\begin{tabular}{|l|c|}
\hline
\textbf{Parameter}                     & \textbf{Value}              \\ \hline
mode                                   & (1) overlay                 \\ \hline
mask mode                              & (4) learned-prd*learned-dst \\ \hline
erode mask modifier                    & 20                          \\ \hline
blur mask modifier                     & 80                          \\ \hline
motion blur power                      & 0                           \\ \hline
output face scale modifier             & 0                           \\ \hline
color transfer to predicted face       & rct                         \\ \hline
sharpen mode                           & (0) None                    \\ \hline
super resolution power                 & 0                           \\ \hline
image degrade by denoise power         & 0                           \\ \hline
image degrade by bicubic rescale power & 0                           \\ \hline
degrade color power of final image     & 0                           \\ \hline
number of workers                      & 12                          \\ \hline
\end{tabular}

\caption{Merge parameters used to generate our face swap stimuli after training.  Note that the original video clips were 1920x1080 pixels at 24 frames per second and faces were extracted at 512x512 pixels.}
\label{tab:merge_params}
\end{table}

\subsection{Analyzing Distribution of Gaze}

\begin{figure}[h]
    \centering
    \includegraphics[width=1\linewidth]{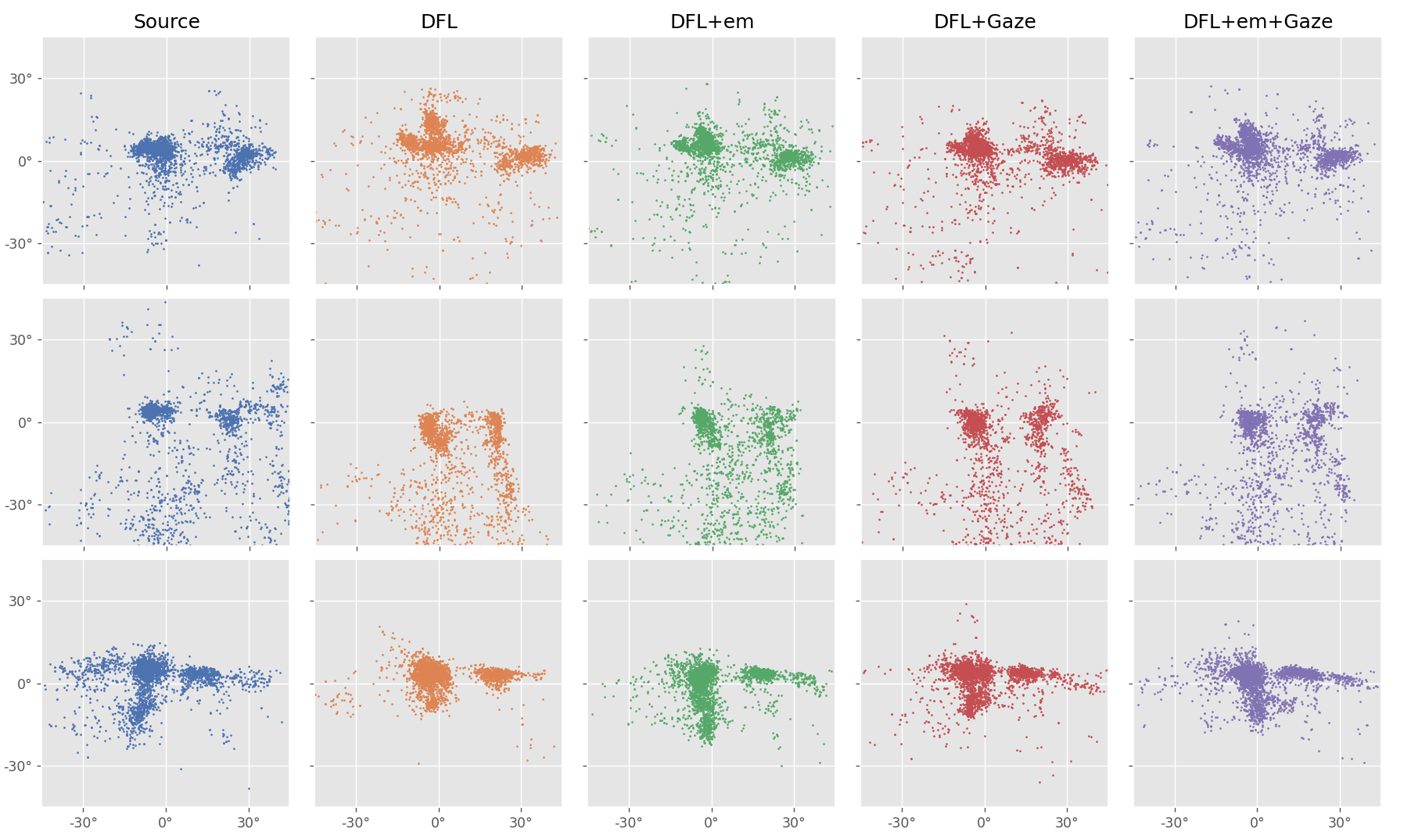}
    \caption{Gaze vectors plotted over all frames of three videos from our dataset (each row corresponding to one full video).  Pitch angles plotted on the horizontal axis and yaw on the vertical axis.}
    \label{fig:gaze_dist}
\end{figure}

We hypothesized that although the DFL+em and DFL+Gaze conditions had similar errors compared to the baseline, the differing approaches may distribute the data in differing manners.  By plotting the pitch and yaw vectors across the frames of video segments (Figure~\ref{fig:gaze_dist}), we see some differences across conditions.  It is visually clear that the baseline DFL distribution does not match the source, and appears to be much more aligned to the horizontal and vertical axes than all other conditions.  The other conditions are much closer in appearance, yet appear to have variations both in outlier distribution and cluster shapes.  For example, the main cluster for DFL+em in the bottom row appears to match the source best along the yaw axis.  However, DFL+Gaze better matches pitch and best mimics the left peninsula that juts out from the bottom of the cluster.

